\documentclass[a4paper]{article}

\usepackage{INTERSPEECH2015}

\usepackage{amsmath,amssymb,bm,graphicx,subfig,float,afterpage,textcomp}
\usepackage[utf8]{inputenc}
\usepackage{pgf, tikz}
\usetikzlibrary{arrows, fit, calc, shapes.geometric,positioning, automata, petri, topaths}

\ninept

\title{Fast and Accurate Recurrent Neural Network Acoustic Models for Speech Recognition}

\makeatletter
\def\name#1{\gdef\@name{#1\\}}
\makeatother
\name{Haşim Sak, Andrew Senior, Kanishka Rao, Françoise Beaufays}
\address{Google\\
{\small \tt \{hasim,andrewsenior,kanishkarao,fsb\}@google.com}}

\begin{document}

\maketitle

\begin{abstract}
We have recently shown that deep Long Short-Term Memory (LSTM) recurrent neural
networks (RNNs) outperform feed forward deep neural networks (DNNs) as acoustic
models for speech recognition.  More recently, we have shown that the
performance of sequence trained context dependent (CD) hidden Markov model
(HMM) acoustic models using such LSTM RNNs can be equaled by sequence trained
phone models initialized with connectionist temporal classification (CTC).  In
this paper, we present techniques that further improve performance of LSTM RNN
acoustic models for large vocabulary speech recognition.  We show that frame
stacking and reduced frame rate lead to more accurate models and faster
decoding.  CD phone modeling leads to further improvements.  We also present
initial results for LSTM RNN models outputting words directly.
\end{abstract}


\section{Introduction}
\label{sec:introduction}

While speech recognition systems using recurrent and feed-forward neural
networks have been around for more than two decades~\cite{Robinson:1996,Bourlard:94}, it
is only recently that they have displaced Gaussian mixture models (GMMs) as the
state-of-the-art acoustic model. More recently, it has been shown that
recurrent neural networks can outperform feed-forward networks on large-scale
speech recognition tasks~\cite{Sak:14a,Sak:14b}. 

Conventional speech systems use cross-entropy training with HMM CD state
targets followed by sequence training. CTC models use a ``blank'' symbol
between phonetic labels and propose an alternative loss to conventional
cross-entropy training. We recently showed
that RNNs for LVCSR trained with CTC can be improved with the sMBR sequence
training criterion and approaches state-of-the-art \cite{Sak:15}.
In this paper we further investigate the use of sMBR-trained CTC models for
acoustic speech recognition and show that with appropriate features and the
introduction of context dependent phone models they outperform the conventional
LSTM RNN models by 8\% relative in recognition accuracy.
The next section describes the LSTM RNNs and summarizes the CTC method and
sequence training. We then describe acoustic frame stacking as well as context
dependent phone and whole-word modeling.
The following section describes our experiments and presents results which are
summarized in the conclusions.

\begin{figure}[!hbtp]
\begin{center}
\includegraphics[width=1.0\hsize]{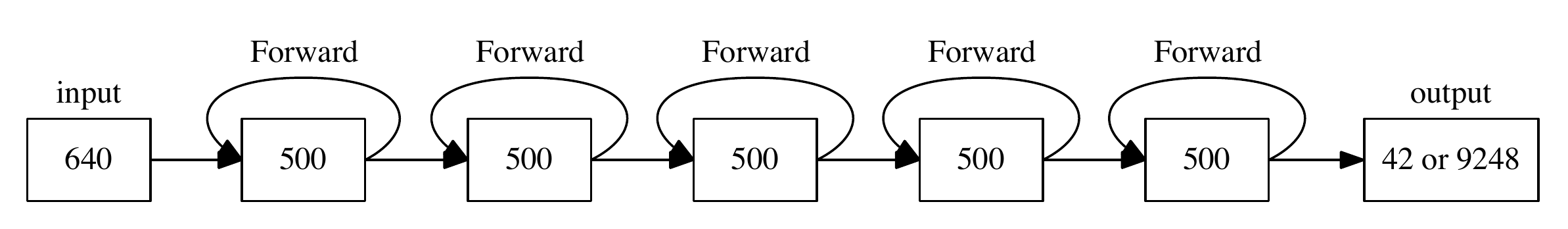} \\
\includegraphics[width=1.0\hsize]{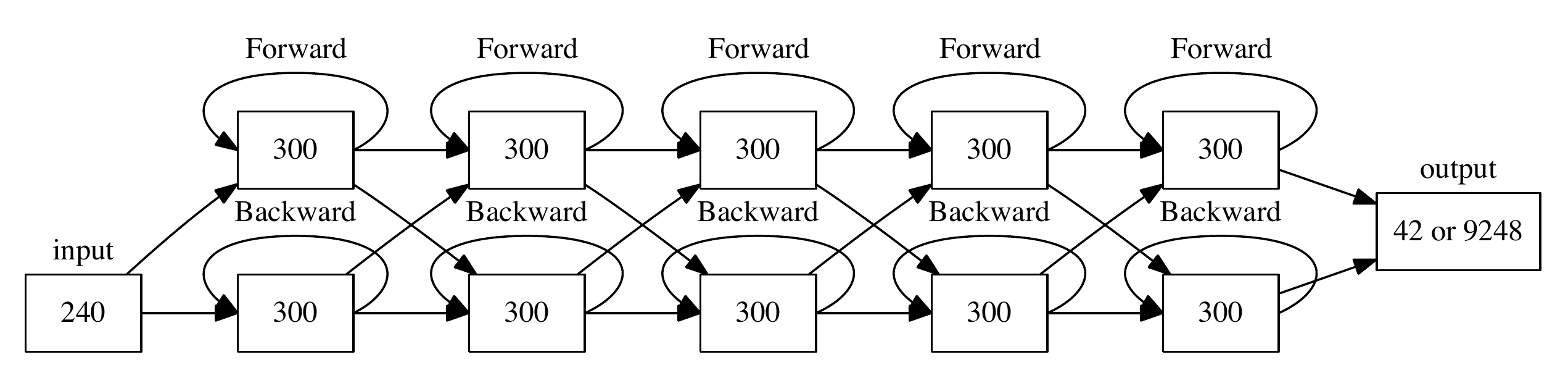}
\end{center}
\caption{Layer connections in unidirectional (top) and bidirectional (bottom)
  5-layer LSTM RNNs.}
\label{fig:lstmrnns}
\end{figure}
\section{RNN Acoustic Modeling Techniques}
\label{sec:methods}
In this work we focus on the LSTM RNN architecture which has shown good
performance in our previous research, outperforming deep neural networks.

RNNs model the input sequence either unidirectionally or bidirectionally
~\cite{Schuster:97}.  Unidirectional RNNs (Figure~\ref{fig:lstmrnns} top) estimate the label posteriors $y_l^t
= p(l_t | x_t, \overrightarrow{h_t})$ using only left context of the
current input $x_t$ by processing the input from left to right and having a
hidden state $\overrightarrow{h_t}$ in the forward direction.  This is
desirable for applications requiring low latency between inputs and
corresponding outputs.
Usually output targets are delayed with respect to features, giving access to a
small amount of right/future context, improving classification accuracy without
incurring much latency.

If one can
afford the latency of seeing the entire sequence, bidirectional RNNs (Figure~\ref{fig:lstmrnns} bottom) 
estimate the label posteriors $p(l_t | x_t, \overrightarrow{h_t},
\overleftarrow{h_t})$ using separate layers for processing the input in the
forward and backward directions.  We use deep LSTM RNN architectures
built by stacking multiple LSTM layers. These have been shown to perform better
than shallow models for speech recognition~\cite{Eyben:09, Graves:13,
  Graves:2013, Sak:14a}.  For bidirectional models, we use two LSTM layers at
each depth --- one operating in the forward and another operating in the
backward direction over the input sequence. Both of these layers are connected
to both the previous forward and backward layers.  The output layer is also
connected to both of the final forward and backward layers.   We experiment with
different acoustic units for the output layer, including context dependent HMM 
states and phones, both context independent and context dependent 
(Section~\ref{sec:cdphones}).  We
train the models in a distributed manner using asynchronous stochastic gradient
descent (ASGD) optimization technique allowing parallelization of training over
a large number of machines on a cluster and enabling large scale training of
neural networks~\cite{Le:2012,Dean:2012,Su:13,Heigold:2013,Sak:14a}.
The weights in all the networks are randomly
initialized with a uniform (-0.04, 0.04) distribution.  We clip the activations
of memory cells to [-50, 50], and their gradients to [-1, 1], making
 CTC training stable.

\subsection{CTC Training}
The CTC approach~\cite{Graves:2006} is
a  technique for sequence labeling using RNNs where the alignment
between the inputs and target labels is unknown.  CTC can be implemented with a
softmax output layer using an additional unit for the \textit{blank} label used
to estimate the probability of outputting no label at a given time.
``Blank'' is similar to the ``non-perceiving state'' proposed earlier~\cite{Morgan:94}.
The output label probabilities from the network define a probability
distribution over all possible labelings of input sequences including the blank
labels.  The network can be trained to optimize the total log probability of
correct labelings for training data as estimated using the network outputs and
forward-backward algorithm~\cite{Rabiner:1989}.
The correct labelings for an input sequence are defined as the set of all
possible labelings of the input with the target labels in the correct sequence
possibly with repetitions and with blank labels permitted between separate
labels.
The targets for CTC training can be efficiently and easily computed using finite state transducers (FSTs) as described in~\cite{Sak:15}, with additional optional blank states interposed between the states of the sequence labels.

While conventional hybrid speech and handwriting recognition systems usually
train from fixed alignments, the use of the forward-backward algorithm to
reestimate network targets given the current model can equally be applied to
conventional recurrent~\cite{Senior:94} or feed-forward
networks~\cite{Senior:14} if no such alignment is available. These conventional
realignment systems have followed the practice of choosing alignments to
maximize the likelihood of the data under state sequence(s) that match the
transcript, and use posteriors scaled by the label priors.

Hence, CTC differs from conventional modeling in two
ways.  First, the additional \textit{blank} label relieves the network from
making label predictions at a frame when it is uncertain.  Second, the training
criterion optimizes the log probability of state sequences rather than the log
likelihood of inputs.

Whether using CTC with posteriors and a blank symbol or a conventional model
with scaled posteriors, once the target posteriors are computed by the
forward-backward algorithm, gradients of the Cross Entropy loss between the
softmax outputs and the targets are backpropagated through the network.

As described in~\cite{Sak:15}, one can use the standard beam search algorithm
for speech decoding with CTC models, again allowing an optional \textit{blank}
state labels between the output labels in the search graph.  In decoding, we
only scale the \text{blank} label posterior by a constant scalar decided by
cross-validation on a held-out set.  We found that CTC models with phone labels
do not require a language model weight to normalize acoustic model scores with
respect to language model scores.  However, CTC models with CD
phone labels (Section~\ref{sec:cdphones}) perform better with a 
weighting constant (2.1).

\subsection{Sequence Discriminative Training}
\label{sec:seqtrain}
Cross-entropy and CTC criteria are suboptimal for the objective of word error rate
(WER) minimization in ASR.  A number of sequence-level discriminative training
criteria incorporating the lexical and language model constraints used in
speech decoding have been shown to improve the performance of DNN and RNN
acoustic models bootstrapped with CE
~\cite{Kingsbury:2009,Kingsbury:2012,Su:13,Povey:2013,Sak:14b} or CTC training
criteria~\cite{Sak:15}.  In this paper, we use the state-level minimum Bayes
risk (sMBR) sequence discriminative training criterion~\cite{Kingsbury:2009}
for improving accuracy of RNN acoustic models initialized with CE or CTC
criterion.  As discussed above and before~\cite{Sak:15}, decoding with CTC
models requires scaling the \textit{blank} label posterior.  We found that sMBR
training can fix this scaling issue if we do not scale the \textit{blank} label
posterior while decoding an utterance to get numerator and denominator lattices
during sMBR training.  Alternatively, the \textit{blank} label scaling can be
baked into into the bias of the \textit{blank} label output unit in the RNN
model by adding negative log of the scale before starting sMBR training, just
as the state priors can be baked into the softmax biases of conventional models
before sequence training. 

To summarize, after sequence discriminative training, the only difference
between CTC and ``conventional'' models is the use of the blank
symbol. Henceforth we use ``CTC'' to refer to these models (and their initial
training using unscaled posteriors to generate alignments) and contrast them
with ``conventional'' models which have no blank symbol, and which, in this
paper, we train with fixed hard alignments.

\vspace{-2mm}\subsection{Acoustic Features}
\label{sec:features}
We use 80-dimensional log mel filterbank energy features computed every 10ms on
25ms windows.  We obtained significant improvements by increasing the number of
filterbanks from 40 up to 80, but only present results for the latter.

\begin{figure}[!hbtp]
\begin{center}
\includegraphics[width=0.75\hsize]{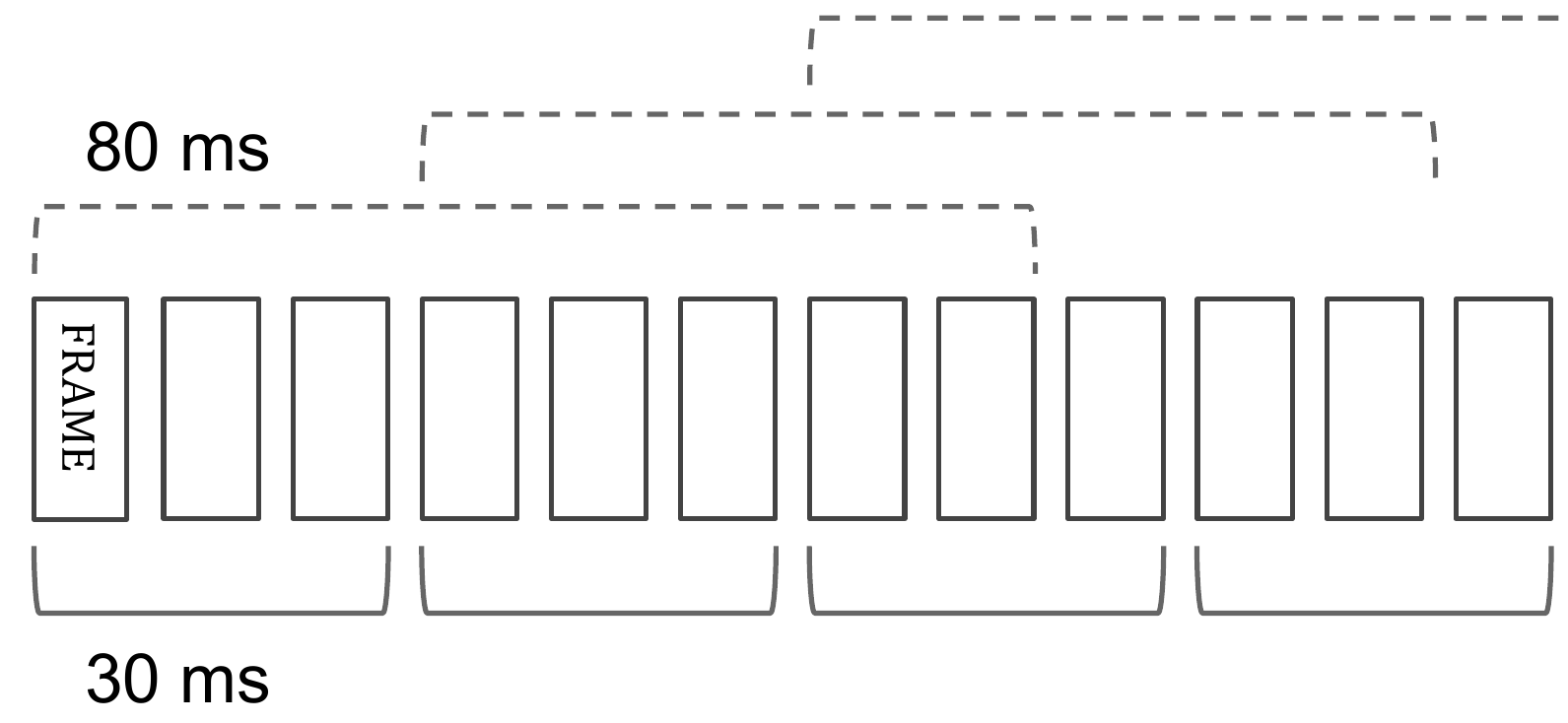}
\end{center}
\caption{Stacking and subsampling of frames. Acoustic features are generated every 10ms, but are concatenated and down-sampled for input to the network: 8 frames are stacked for unidirectional (top) and 3 for bidirectional models (bottom).}
\label{fig:framestacking}
\end{figure}

In the past, we have observed that training with CTC is unstable, in that some
training runs fail to converge. We found~\cite{Sak:15} that stability was
improved by starting training using two output layers with CTC and the
conventional CE loss, or initializing from a network whose LSTM layers had been
pretrained using the CE loss.  We suggest that this is because of the inherent
arbitrariness of the alignment with CTC, which considers valid any alignment in
which the target symbols are emitted in the correct order interspersed with an
arbitrary number of blanks. One way of reducing the huge space of alignments is
to reduce the number of input frames. This can be done by simply decimating the
input frames, though to present the full acoustic information of the input
signal, we first stack frames so that the networks sees multiple (e.g. 8) frames at
a time but then decimate the frames so that we skip forward multiple frames (e.g. 3) after
processing each such ``super-frame''. This process is illustrated in Figure~\ref{fig:framestacking}.

By decimating the frames in this manner, the acoustic model is able to process
the full signal but acoustic model computation need only happen every 30ms. For
a network of a fixed size this results in a dramatic reduction in the acoustic
model computation and decoding time.
\vspace{-1mm}
\vspace{-2mm}\subsection{Context-Dependent Phones}
\label{sec:cdphones}
Previously CTC models \cite{Graves:13,Sak:15} used context
independent outputs, yet it is well known that context dependent states
outperform context independent models for conventional speech recognition
systems, both GMM-based and neural-network hybrids.  We argue that context
dependency is an important constraint on decoding and provides a
useful labeling for state outputs, so believe it should be useful for CTC
models.

Previously, \cite{Senior:15} it was shown that it is possible to build
context dependent whole-phone models, and that for LSTM-HMM hybrid speech
recognition, these models can give similar results to context dependent state
models, provided that a minimum duration is enforced. We repeat that procedure,
using the hierarchical binary divisive clustering algorithm of Young {\em et
  al.}~\cite{YoungOW94} for context-tying. We use three frames of
40-dimensional log-mel filterbanks to represent each whole-phone instance. One
tree per phone is constructed, with the maximum-likelihood-gain phonetic
question being used to split the data at each node.  On our training data we
end up with 9287 CD phones.

As found before, enforcing a minimum duration for each phone was found to
improve word error rates, and we again use a 10\% cutoff of the training-set
duration histogram as the minimum duration for each CD-phone for decoding of
our conventional models.  For CTC, no such duration model is imposed.
\vspace{-4mm}\subsection{Word Acoustic Models}
The combination of LSTM RNNs' memory and CTC's ability to learn an alignment
between label and acoustic frame sequences, while relieving the network from
having to label each frame by introducing the \textit{blank} label, enables the
use of longer duration modeling units.  For instance, we can train
acoustic models predicting whole words rather than
phonemes. There have been previous studies using LSTM RNN CTC models for
keyword spotting tasks with small vocabularies (e.g. 12 words~\cite{Fernandez:07}).
In this paper, we
investigate the effectiveness of word acoustic models trained over a large
training set with various large vocabularies ranging from 7,000 to 90,000
words.
\vspace{-2mm}\section{Experiments}
\label{sec:experiments}
\vspace{-2mm}\subsection{Data \& Models}
\label{sec:eval}We train and evaluate LSTM RNN acoustic models on hand-transcribed,
anonymized utterances taken from real 16kHz Google voice search traffic.  Our
training set consists of 3 million utterances with average duration of about
4s.  To achieve robustness to background noise and reverberant environments we
synthetically distort each utterance in a room simulator with a virtual noise
source. Noise is taken from the audio of YouTube videos. Each utterance is
randomly distorted to get 20 variations. This ``Multi-style training'' also
alleviates overfitting of CTC models to training data.

The test set's 28,000 utterances are each distorted once with held-out noise
samples.  Evaluation uses a 5-gram language model pruned to 100 million
$n$-grams.  Rescoring word lattices with a larger $n$-gram model gives similar
relative gains for all the acoustic models, therefore we only report results
after first pass decoding.  For all the experiments, we use a wide beam in
decoding to avoid search errors and obtain the best possible performance.

For training networks with CE criterion using fixed
alignments, the training utterances are force-aligned using an 85 million
parameter DNN with 13522 CD HMM states.  We explored variations of frame
stacking and skipping as described in section~\ref{sec:features}. For the
conventional unidirectional models' inputs, we either stack 8
consecutive feature frames and skip 1 frame or present a single frame with a 5
frames delayed target --- both approaches give similar results.  For the
bidirectional models, we only need to use a single frame input.
For bidirectional CTC models, we stack 3 consecutive feature frames as input
feature vector and skip 3 frames.  For unidirectional CTC models, we stack 8
consecutive feature frames and skip 3 frames (Figure~\ref{fig:framestacking}).
We found longer context helps unidirectional models but is not needed for
bidirectional models.

For CTC models, we obtained the best results with depth 5. Unidirectional
models used 500 memory cells in each layer and bidirectional models had 300
memory cells for each direction in each layer.  For the conventional models, we
got the best results with 2 LSTM layers of 1000 cells each with a recurrent
projection layer of 512 units.

\subsection{Results \& Discussion}
\label{sec:results}

Table~\ref{tab:wer} shows the word error rates (WERs) on the voice search task
for various unidirectional and bidirectional LSTM RNN acoustic models trained
with CE or CTC loss with CD HMM state, CI phone or CD phone labels.  As can be
expected from trying to learn with 3 state HMM labels, CTC CD state
models do not perform well.  The unidirectional CE CD phone model is
marginally better than the corresponding CE CD state model.  CTC CI phone
models perform very similarly to CE CD state models.  CTC CD phone models give
significant improvements over CTC CI phone models -- about 8\% for
unidirectional and 3.5\% for bidirectional.  Bidirectional models improve over
unidirectional ones about 10\% for CD state and CI phone models -- while
improving CTC CD phone models only 5\%.

\begin{table} [t,h]
\vspace{2mm}
\centerline{
\begin{tabular}{|c|c|c|c|c|}
\hline
Labels & \multicolumn{2}{c|}{CE (\%)} & \multicolumn{2}{c|}{CTC (\%)}\\
\cline{2-5}
       & Uni & Bi & Uni & Bi\\
\hline
\hline
CD state & 15.6 & 14.0 & 18.9 & 16.5 \\
\hline
CI phone & \multicolumn{2}{c|}{} & 15.5 & 14.1 \\
\hline
CD phone & 15.5 &   & 14.3 & 13.6 \\
\hline
\end{tabular}
}
\caption{\label{tab:wer} {\it WERs for conventional and CTC initialization of LSTM RNN acoustic models.}}
\end{table}
Table~\ref{tab:smbr_wer} shows the results for sequence discriminative training
of these initial CE/CTC models with sMBR loss.
We can see that sMBR training consistently improves all of the models initially
trained with CE or CTC loss about 10\% relative.
We obtain best results with CTC CD phone models outperforming the second best model
about 8\% for unidirectional and 4\% for bidirectional model.
\begin{table} [t,h]
\vspace{2mm}
\centerline{
\begin{tabular}{|c|c|c|c|c|c|}
\hline
Labels & \multicolumn{3}{c|}{Initialization} & \multicolumn{2}{c|}{+sMBR}\\
\cline{2-6}
       & Method  & Uni & Bi & Uni & Bi\\
\hline
\hline
CD state & CE & 15.6 & 14.0 & 14.0 & 12.9 \\
\hline
CI phone & CTC & 15.5 & 14.1 & 14.2 & 12.7 \\
\hline
CD phone & CTC & 14.3 & 13.6 & 12.9 & 12.2 \\
\hline
\end{tabular}
}
\caption{\label{tab:smbr_wer} {\it WERs (\%) for sequence-trained LSTM RNN models.}}
\end{table}

\begin{figure}[h]
  \centering
  \subfloat[unidirectional phone CTC + sMBR]{\includegraphics[width=0.45\textwidth]{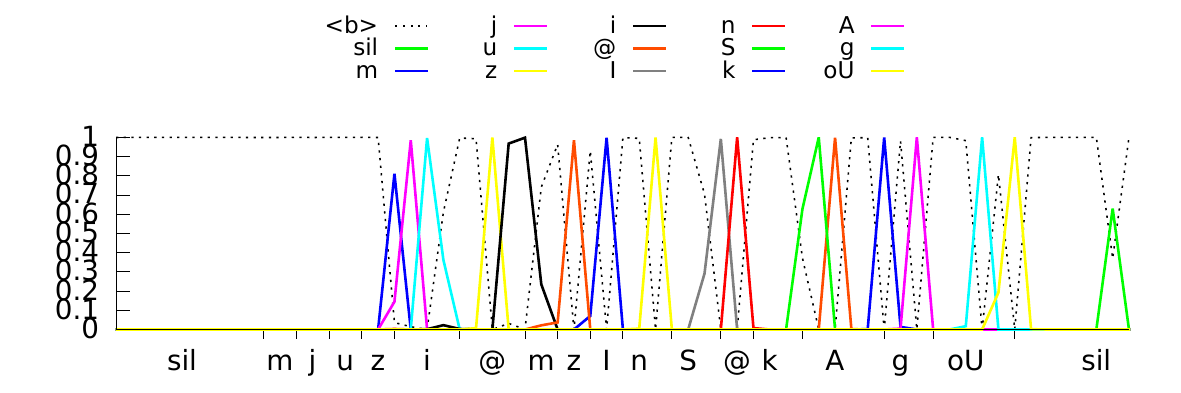}}\\
  \subfloat[unidirectional CD phone CTC]{\includegraphics[width=0.45\textwidth]{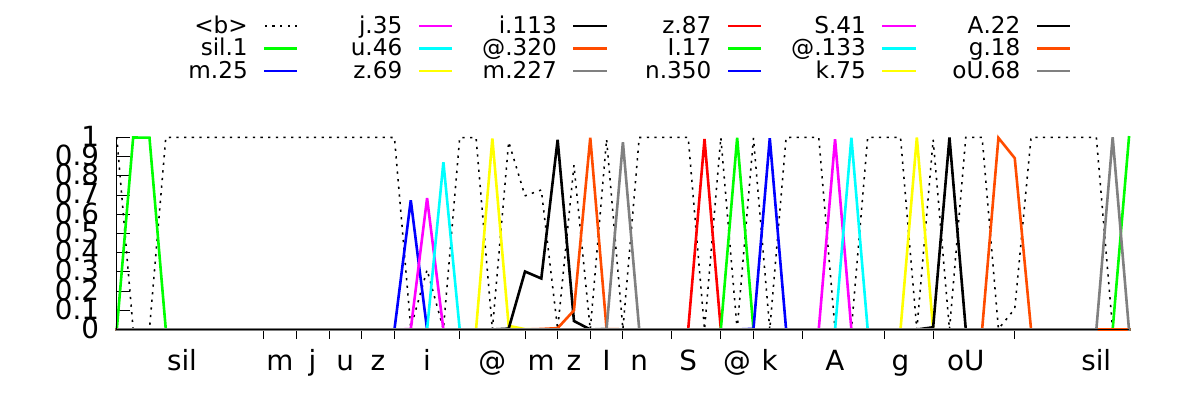}}\\
  \subfloat[unidirectional CD phone CTC + sMBR]{\includegraphics[width=0.45\textwidth]{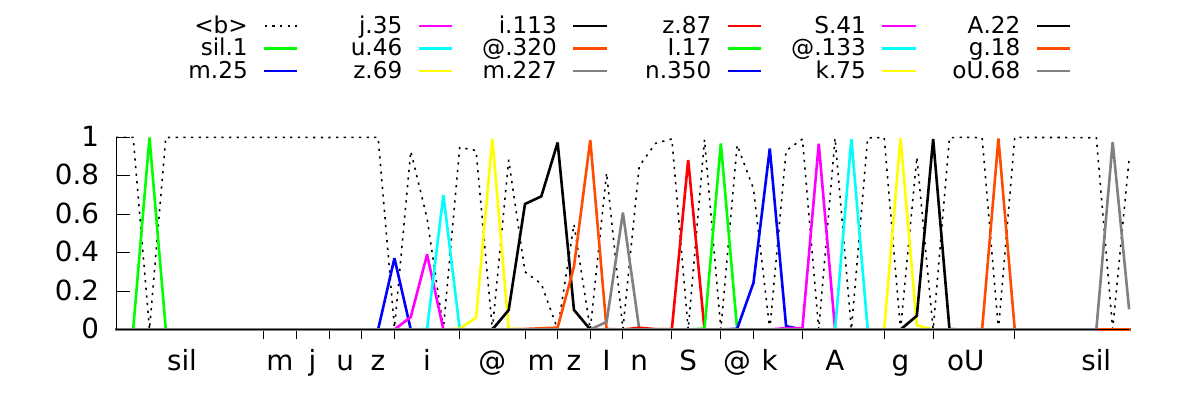}}\\
  \subfloat[bidirectional CD phone CTC + sMBR]{\includegraphics[width=0.45\textwidth]{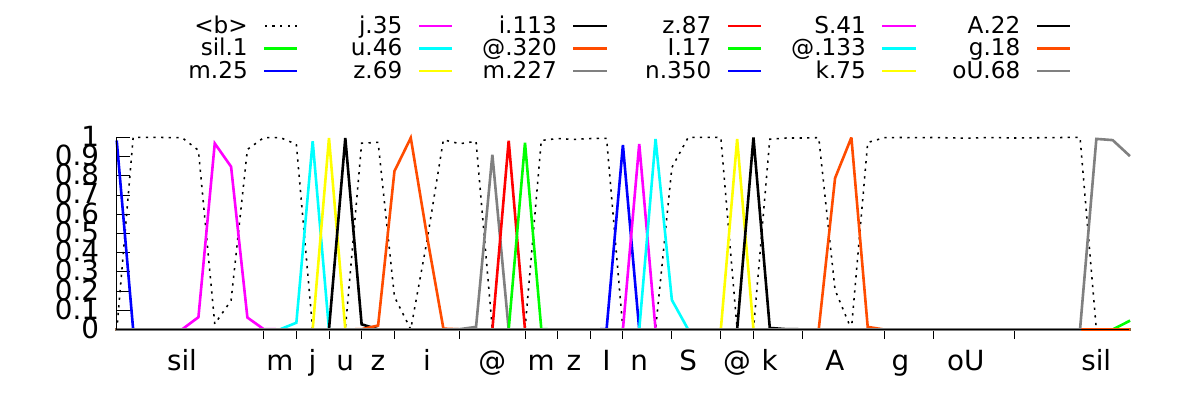}}\\
  \caption{Label posteriors estimated by various LSTM RNN models plotted against fixed DNN frame level alignments shown only for labels in the alignment on a heldout utterance \textit{`museums in Chicago'}. \textit{\textless b\textgreater} refers to the \textit{blank} label.}
  \label{fig:posteriors}
\end{figure}

Figure~\ref{fig:posteriors} shows label posteriors estimated by various CTC phone and CD phone models.
It can be seen that spikes for the label posteriors do not correspond to the DNN alignment and differ between the models.
Unidirectional models delay their output labels by about 300 milliseconds.
As can be expected, bidirectional models make better predictions.
The models are not good at modeling \textit{silence} labels.
Sequence discriminative training changes posteriors, but not the spike positions.

\begin{figure}[h]
  \centering
  \includegraphics[width=0.45\textwidth]{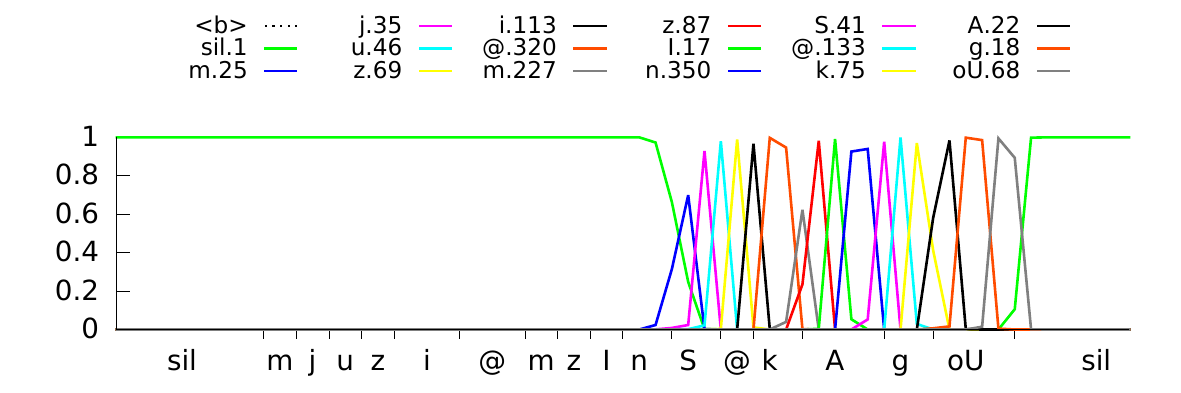}
  \caption{Label posteriors for a unidirectional LSTM RNN model with conventional alignment -- no blank label.}
  \label{fig:conv_align}
\end{figure}

While learning an alignment in conventional GMM-HMM systems and DNN-HMM hybrid
systems has been shown to work well, learning a conventional alignment without
\textit{blank} label using LSTM RNNs does not work well.
Figure~\ref{fig:conv_align} shows label posteriors estimated using a
unidirectional LSTM RNN CD phone model trained with CTC loss with no
\textit{blank} label allowed between CD phone labels. The model has learned an
arbitrary alignment.  Having a memory as in RNN models in contrast to
memoryless feed forward neural networks means the model can delay its outputs instead of making decisions using only local temporal
information. Therefore, the model learns an alignment where it chooses to
adjust its labeling according to its certainty for a label given the input.
This results in an arbitrary alignment where some labels are repeated more than
others depending on the input. Note that using a hybrid approach with a prior
cannot fix this issue.

\begin{figure}[t,h]
  \centering
  \subfloat[7k vocabulary]{\includegraphics[width=0.5\textwidth]{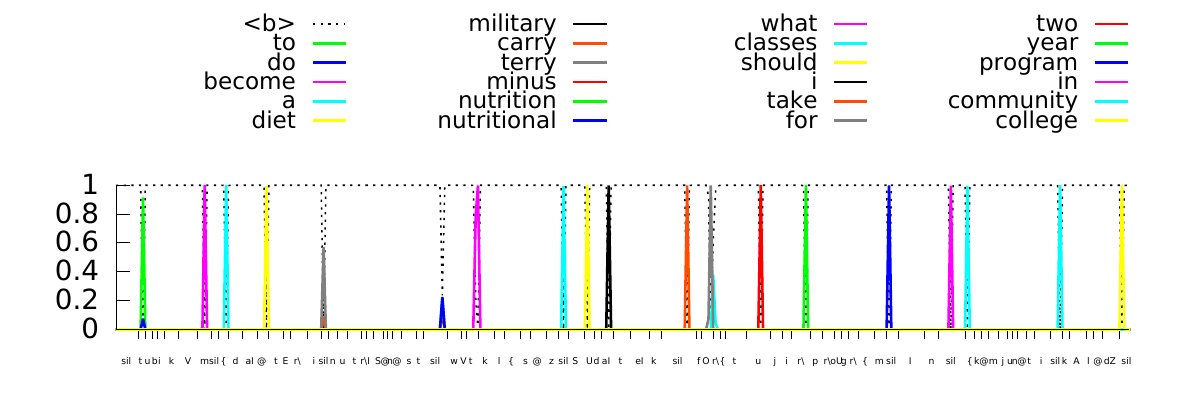}}\\
  \subfloat[90k vocabulary]{\includegraphics[width=0.5\textwidth]{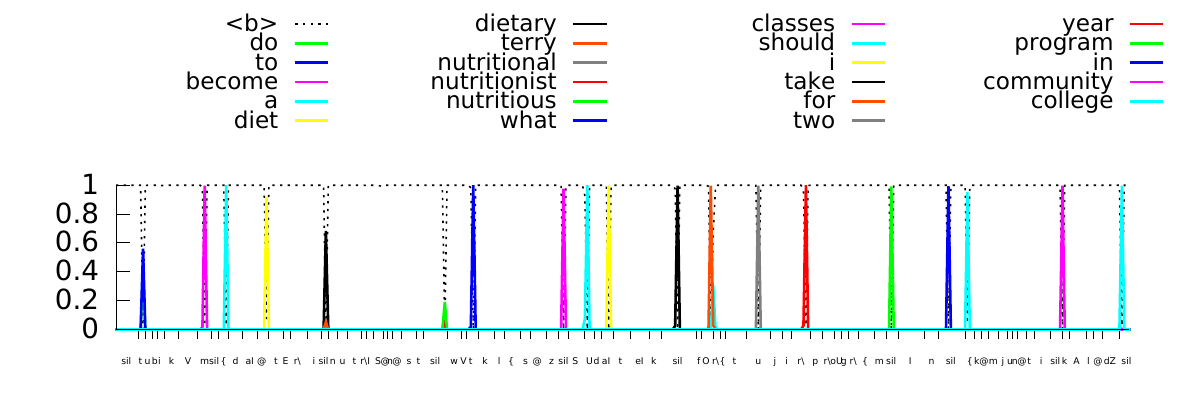}}
  \caption{\textit{`To become a dietary nutritionist what classes should I take for a two year program in a community college'}}
  \label{fig:word}
\end{figure}

We have experimented with CTC word models with various vocabulary sizes where
the output layer directly predicts words rather than phonemes.  We have used
two different vocabularies with the most frequent 7,000 ($\geq 150$ exemplars)
or 25,000 ($\geq 20$ exemplars) words from the training data transcripts.
Table~\ref{tab:wer_word} shows WERs for bidirectional CTC word acoustic models
as calculated edit distance between reference word sequence and predicted word
sequence where the word with highest probability is taken ignoring repetitions
and the \textit{blank} label with no language model or decoding.  We have also
experimented with 90k vocabulary CTC word models, and note that the bidirectional
model gives a 25\% lower WER than the unidirectional model.
\begin{table} [t,h]
\vspace{2mm}
\centerline{
\begin{tabular}{|c|c|c|c|c|}
\hline
Vocabulary & OOV & WER (\%) & In vocab. WER (\%) \\
\hline
\hline
25k Word & 4.8 & 19.5 & 14.5 \\
\hline
7k Word & 13 & 26.8 & 11.8 \\
\hline
\end{tabular}
}
\caption{\label{tab:wer_word} {\it LSTM RNN CTC word acoustic models.
The WERs and out of vocabulary (OOV) rates for word models are on heldout data with no decoding or language model.
WERs in the last column are computed ignoring utterances containing OOVs.}}
\end{table}
Figure~\ref{fig:word} shows label posteriors estimated by the bidirectional CTC models with 7k
and 90k vocabulary for a heldout utterance. We plot the posteriors for all the labels
that were above 0.05 probability at any time.
The words 'dietary' and 'nutritionist' are OOV for the 7k vocabulary.
It is interesting to see that the models make spiky predictions even with a large vocabulary
and the predictions for confused words are output at the same time.
Although these two models have very similar spike positions for the words,
they are different for the 25k model.

\vspace{-4mm}\section{Conclusions}
\label{sec:conclusions}
In this work we have shown a number of improvements to recurrent network
acoustic models. The use of longer-term feature representations, processed at
lower frame rates brought stability to the convergence of CTC training of
models with blank symbol outputs while also resulting in a considerable
reduction in computation. After sequence training, such models are found to
perform better than previous acoustic models. Performance of the blank-symbol
acoustic models was further improved by the introduction of context-dependent
phonetic units, with the result that these models now outperform conventional
sequence trained LSTM-hybrid models.
We have also shown that we can train word level acoustic models to achieve reasonable accuracy
on medium vocabulary speech recognition without using a language model.

\clearpage
\newpage
\eightpt
\bibliographystyle{IEEEtran}
\bibliography{ctc}

\end{document}